\documentclass{article}
\usepackage[utf8]{inputenc}
\usepackage{graphicx}
\usepackage{natbib}
\usepackage{hyperref}
\usepackage{amsmath}
\usepackage{amssymb}
\usepackage{url}
\usepackage{tikz}
\usepackage{makecell}
\usepackage{pifont}
\usepackage{etoolbox}
\newcommand{\cmark}{\ding{51}}  
\newcommand{\xmark}{\ding{55}}  
\usepackage{makecell}
\usepackage{amssymb}  
\usetikzlibrary{arrows.meta, positioning}

\title{\textbf{The STROT Framework: Structured Prompting and Feedback-Guided Reasoning with LLMs for Data Interpretation}
 }
\author{
  Amit Rath
  \texttt{ ar4759@nyu.edu}
}

\begin{document}

\maketitle

\begin{abstract}
Large language models (LLMs) have demonstrated remarkable capabilities in natural language understanding and task generalization. However, their application to structured data analysis remains fragile due to inconsistencies in schema interpretation, misalignment between user intent and model output, and limited mechanisms for self-correction when failures occur. 

This paper introduces the \textbf{STROT Framework} (Structured Task Reasoning and Output Transformation), a method for structured prompting and feedback-driven transformation logic generation aimed at improving the reliability and semantic alignment of LLM-based analytical workflows. STROT begins with lightweight schema introspection and sample-based field classification, enabling dynamic context construction that captures both the structure and statistical profile of the input data. This contextual information is embedded in structured prompts that guide the language model toward generating task-specific, interpretable outputs.

To address common failure modes in complex queries, this framework incorporates a refinement mechanism in which the model iteratively revises its outputs based on execution feedback and validation signals. Unlike conventional approaches that rely on static prompt templates or single-shot inference, and treats the LLM as a reasoning agent embedded within a controlled analysis loop—capable of adjusting its output trajectory through planning and correction. 

The result is a robust and reproducible framework for reasoning over structured data with LLMs, applicable to diverse data exploration and analysis tasks where interpretability, stability, and correctness are essential.
\end{abstract}

\section{Introduction}

Large language models (LLMs), including widely adopted systems such as GPT-3.5, GPT-4, Claude, and open-source alternatives like Mistral and Mixtral, have demonstrated considerable generalization capabilities across a wide spectrum of natural language tasks. Pretrained on large-scale corpora and instruction-tuned for downstream usability, these models are increasingly relied upon for tasks such as text summarization, question answering, transformation logic generation, and conversational reasoning. Their ability to process natural language instructions in both zero-shot and few-shot settings has enabled the development of user-facing tools for knowledge retrieval, dialogue, and programmatic interaction.

Despite these advances, LLMs remain unreliable when tasked with analytical reasoning over structured data representations—such as tabular datasets, relational outputs, or semantically typed records. In such settings, models frequently hallucinate field names, conflate categorical and numerical columns, or produce outputs that are syntactically valid but semantically incoherent. While models like GPT-4 and Claude show measurable improvements in transformation logic synthesis and chain-of-thought reasoning, they continue to underperform when exposed to unfamiliar schemas, domain-specific naming conventions, or multi-step transformation requirements. These issues are particularly pronounced in open-ended exploratory tasks, where schema comprehension, iterative analysis, and user intent alignment are essential.

These limitations stem from a structural mismatch between the training objectives of LLMs and the demands of structured reasoning. Whereas many natural language applications are tolerant of ambiguity and variation, analytical systems require stability under schema perturbation, interpretability of intermediate steps, and robustness to runtime failures. Existing prompt-based systems tend to treat LLMs as static translators—capable of emitting a final answer in one pass—without incorporating mechanisms for structured planning or post-hoc correction. As a result, current approaches are often brittle, non-reproducible, and unsuitable for tasks that require fine-grained reasoning over structured inputs.

\subsection*{Problem Statement}

Current approaches to LLM-based structured data analysis often rely on a single-pass interaction model: the model is prompted with a flattened representation of the schema and a natural language query, and is expected to generate a complete and executable output in one step. While this method has shown promise across various structured reasoning tasks \citep{zhongSeq2SQL2017, yu2018spider, li2023resdsql}, it introduces limitations in both robustness and interpretability.

Specifically, such one-shot prompting models frequently lack explicit schema grounding, leading to hallucinated field references or misaligned field usage. Their outputs tend to be highly sensitive to prompt phrasing, often yielding divergent results for semantically equivalent inputs. Moreover, these systems typically do not provide structured avenues for recovering from failure—whether due to execution errors, semantic mismatches, or partial reasoning—thereby limiting their reliability in production-grade analytical workflows.

These challenges highlight a broader gap in the design of LLM-driven data tools: the absence of intermediate reasoning steps, contextual anchoring, and iterative feedback mechanisms. In high-stakes domains that demand accuracy, transparency, and semantic control, such capabilities are essential for bridging the gap between natural language and structured output.

\subsection*{Motivation}

Human analysts do not approach structured data analysis as a single-step task. Rather, they begin by inspecting available fields,   identifying relevant dimensions, interpreting distributions, and forming preliminary hypotheses. The analysis process is inherently iterative: analysts validate assumptions through trial and error, revise their logic in response to intermediate results, and adapt their approach based on evolving insights and context.

In contrast, current LLM-based systems typically rely on rigid, single-shot prompting mechanisms that fail to replicate this adaptive reasoning pattern. They lack mechanisms to incorporate schema structure, validate intermediate reasoning steps, or recover from execution failures. As a result, they often produce brittle or semantically inconsistent outputs when applied to real-world structured data tasks.

The motivation for this work is to formalize a more robust interaction model—one that mirrors human-like iterative analysis—by embedding the language model within a scaffolded framework that supports schema-aware interpretation, structured planning, and feedback-driven output refinement. This agentic formulation enables the model to function not merely as a passive responder but as a reasoning component capable of adjusting its behavior based on both data structure and task outcomes. By doing so, we seek to bridge the gap between the generality of LLMs and the precision, accountability, and adaptability required for structured data analysis.

\subsection*{Proposed Approach: The STROT Framework}

This paper introduces the  \textbf{S}tructured \textbf{T}ask \textbf{R}easoning and \textbf{O}utput \textbf{T}ransformation, a structured prompting and feedback-driven approach for improving the reliability of LLMs in data exploration and analysis tasks. The framework is built on the premise that structured data understanding requires contextual grounding, semantic alignment, and adaptive behavior in response to execution outcomes.

STROT consists of three core components:

\begin{itemize}
\item \textbf{Schema-Guided Context Construction:} Prior to model invocation, the system performs lightweight schema introspection and sample-based field classification. The resulting context explicitly encode categorical, numerical, and temporal fields, reducing semantic ambiguity and improving input fidelity.

\item \textbf{Goal-Aligned Prompt Scaffolding:} Prompt templates are constructed dynamically based on the analytical goal, data schema, and available samples. This ensures that the model’s reasoning is grounded in the structural and statistical profile of the input, aligning generated logic with user intent.

\item \textbf{Feedback-Based Output Refinement:} Generated outputs are treated as provisional. If execution results in failure, empty output, or semantic mismatch, a structured refinement mechanism is triggered, prompting the model to revise its output based on runtime feedback and correction signals.

\end{itemize}

\section{Related Work}

\subsection*{LLMs for Structured Data Interpretation}

Recent advances in large language models (LLMs) have led to growing interest in their application to structured data analysis tasks. Rather than operating solely on natural language corpora, researchers have begun adapting LLMs to reason over data tables, relational structures, and typed inputs commonly found in enterprise and scientific settings. One line of work investigates the alignment between tabular schemas and natural language prompts, where the goal is to equip the model with structural context such as column types and statistical previews \citep{herzig2020tapas, zhong2022feta}. These methods often rely on prompt engineering or schema serialization, yet they struggle to scale in settings with large or dynamic schemas.

Studies such as \citet{lu2023learn} explore schema-grounded prompting techniques to teach LLMs to extract and transform tabular data, while \citet{liang2023holistic} examine instruction tuning for domain-specific table reasoning. Despite these developments, most approaches treat LLMs as passive mappers from structured context to static outputs, often assuming correctness in a single pass. This limits their robustness in real-world analytical workflows that require error detection, revision, or exploration.

\subsection*{Feedback-Driven and Iterative Refinement Frameworks}

Beyond static prompting, emerging research explores mechanisms for iterative reasoning and output correction. The SELF-REFINE framework proposed by \citet{madaan2023selfrefine} introduces a loop where the model generates a draft output, reflects on it, and improves it based on self-generated critiques. In the domain of program synthesis, approaches such as CoCoGen \citep{wang2023cocogen} integrate execution-based feedback to refine generated transformation logic through multiple iterations, often incorporating compiler errors or test results.

These methods demonstrate the value of treating LLMs as agents capable of adaptive refinement rather than fixed-output generators. However, most existing work focuses on natural language or transformation logic tasks and does not address structured data exploration, where semantic accuracy and alignment with schema constraints are critical.

This framework builds on recent advances in prompt-based reasoning by introducing an agentic interaction loop tailored for structured data tasks. The language model operates within a scaffolded execution environment that combines schema-aware context construction, sample-driven grounding, and iterative prompt refinement. Rather than relying on single-pass inference or post-hoc human correction, the system treats data interpretation as a multi-step, feedback-aware process. Each step—from planning to execution—is mediated by structured prompts, enabling the model to engage in chain-of-thought reasoning, detect failures, and revise its outputs through programmatic retries. This design promotes robustness in high-variance tabular settings, where semantic alignment and execution fidelity are critical.

\subsection*{Comparative Summary}

Table~\ref{tab:method-comparison} summarizes key distinctions between STROT and representative approaches from the structured data reasoning literature. Seq2SQL~\citep{zhongSeq2SQL2017} and ResdSQL~\citep{li2023resdsql} focus on structured query generation using learned schema representations but operate in a single-shot manner and are narrowly scoped to SQL generation tasks. Self-Refine~\citep{madaan2023selfrefine} introduces feedback-driven refinement but lacks schema-specific reasoning or explicit planning structures. In contrast, STROT combines schema typing, scaffolded intermediate planning, and iterative self-correction into a unified prompting framework applicable to a broader range of exploratory and interpretive data tasks. This multi-phase design allows the model to adaptively reason over structured inputs with higher reliability and transparency than traditional one-pass prompting pipelines.

\begin{table}[h]
\centering
\small
\caption{Comparison of STROT with representative methods for structured data interpretation.}
\begin{tabular}{lcccc}
\hline
\textbf{Method} & 
\makecell{\textbf{Schema} \\ \textbf{Awareness}} & 
\makecell{\textbf{Planning} \\ \textbf{Stage}} & 
\makecell{\textbf{Feedback} \\ \textbf{Loop}} & 
\makecell{\textbf{Task} \\ \textbf{Generality}} \\
\hline
Seq2SQL \citep{zhongSeq2SQL2017}         & \cmark & \xmark & \xmark & \xmark \\
ResdSQL \citep{li2023resdsql}            & \cmark & \xmark & \xmark & \xmark \\
Self-Refine \citep{madaan2023selfrefine} & \xmark & \xmark & \cmark & \cmark \\
\textbf{STROT}                           & \cmark & \cmark & \cmark & \cmark \\
\hline
\end{tabular}
\end{table}

\section{Methodology}

The STROT Framework (Structured Task Reasoning and Output Transformation) is designed to enable large language models (LLMs) to function as autonomous reasoning agents for automated exploration and interpretation of structured tabular data. Traditional approaches to leveraging LLMs for data-related tasks commonly adopt a single-pass prompt-response paradigm, where a natural language instruction is translated into a static output—such as a query, transformation, or transformation logic snippet—in one inference step. This mode of operation assumes that the model can simultaneously understand the data schema, correctly infer the user’s intent, and produce a semantically and syntactically valid solution without external feedback or internal revision.

It departs fundamentally from this assumption by embedding the LLM within a multi-phase, feedback-driven pipeline that treats data understanding as a dynamic and structured process. Rather than issuing one-shot completions, the framework decomposes the broader task of data interpretation into sequential stages: schema introspection, task planning, transformation logic synthesis, and error-aware refinement. Each stage is explicitly structured and conditioned on intermediate outputs, allowing the model to incrementally build and verify its reasoning.

At the core of STROT is the notion of the LLM as a structured reasoning component—one that can analyze data context, generate interpretable plans, synthesize transformation logic, and respond to runtime signals in an iterative fashion. The framework is model-agnostic and modular: it does not rely on fixed prompt templates, handcrafted rules, or fine-tuned parameters. Instead, it leverages generic LLM capabilities through carefully scaffolded prompts and controlled execution loops. This architecture enables to maintain robustness in the face of schema variability, ambiguous instructions, or partial failure, ultimately yielding a more resilient and interpretable process for automated structured data analysis.

\subsection{Architecture Overview}

The overall workflow of this framework is composed of three interdependent components, each responsible for a distinct phase of the reasoning process. Together, these components enable a large language model to operate not as a one-shot predictor, but as a structured reasoning agent capable of interpreting and transforming tabular data through modular, iterative decision-making.

\begin{enumerate}
    \item \textbf{Schema-Aware Context Construction:} The first stage involves analyzing the underlying dataset to extract a structured and interpretable representation of its schema. This includes determining the semantic types of columns (e.g., categorical, numerical, temporal), summarizing statistical properties such as value distributions and cardinality, and selecting representative data samples. The goal is to construct a schema context $\mathcal{C}$ that conveys sufficient information about the structure and content of the dataset in a form that can be passed to the language model. By grounding the model in a well-defined context, this step reduces ambiguity, mitigates the risk of field hallucination, and improves alignment between model behavior and the actual data.

    \item \textbf{Prompt Scaffolding and Task Planning:} In the second phase, the system translates the user’s natural language intent $\mathcal{Q}$ into a structured prompt that combines both the schema context $\mathcal{C}$ and the task specification. This prompt is then used to elicit from the language model a high-level analysis plan $\mathcal{P}$ that outlines how the data should be interpreted or transformed. The plan typically includes a description of the logical steps required, the specific fields involved, and the type of transformation or reasoning to be applied. This intermediate representation serves as a bridge between human intent and executable logic, enabling the system to validate and control the model’s reasoning prior to any actual data processing.

    \item \textbf{Feedback-Driven transformation logic Synthesis and Refinement:} In the final stage, the analysis plan $\mathcal{P}$ is translated into executable logic — typically in the form of a function or program—that can be applied to the input dataset. The logic is executed in a controlled runtime environment, and the output is validated against structural and semantic expectations. If the execution fails due to logical errors, invalid operations, or unmet constraints, the system enters a feedback loop wherein the language model is asked to revise the logic based on the error trace or result analysis. This process continues iteratively until either a valid output is produced or a maximum number of refinement attempts is reached. By treating execution failures as recoverable and instructive, this component allows the system to improve robustness and emulate aspects of human-like troubleshooting and revision.
\end{enumerate}

Each of these stages plays a critical role in ensuring the reliability, interpretability, and adaptability of LLM-based structured data reasoning. The modular design allows to generalize across data domains and model architectures, while the feedback-oriented execution loop offers a principled alternative to static prompting.

We now describe each component in detail.

\subsection{Schema-Aware Context Construction}

Let $\mathcal{D} = \{r_1, \dots, r_n\}$ denote a structured tabular dataset with $n$ rows and $m$ columns, where each row $r_i$ is a tuple defined over the feature set $\{c_1, \dots, c_m\}$. The goal of this stage is to construct a semantically rich yet token-efficient representation of the dataset schema that can be reliably consumed by a large language model (LLM) within the constraints of its context window.

For each column $c_j$ $(1 \le j \le m)$, a profiling function $\phi : c_j \mapsto \tau_j$ is used to infer a high-level semantic type $\tau_j$, based on both syntactic patterns and empirical value distributions:

\[
\tau_j =
\begin{cases}
\text{numerical} & \text{if } c_j \text{ contains quantitative values with high cardinality}, \\\\
\text{temporal} & \text{if } c_j \text{ conforms to known date/time formats}, \\\\
\text{categorical} & \text{otherwise}.
\end{cases}
\]

To support grounding and downstream task planning, the system additionally computes a compact statistical signature $S_j$ for each column, which includes relevant features such as:

\[
S_j = \left(\text{cardinality},\ \text{null rate},\ \text{value range or bounds},\ \text{distributional skew or entropy } \right)
\]

This metadata provides the LLM with coarse-grained priors over the structure of the data and helps constrain the space of plausible operations. To further aid the model’s ability to align prompts with data semantics, a sample of $k$ representative cell values is extracted from each column using either uniform or stratified sampling heuristics:

\[
\{r_i[c_j]\}_{i=1}^k \subseteq \mathcal{D}
\]

These samples function similarly to in-context examples in few-shot prompting, offering concrete instantiations of the column's domain without requiring full-row serialization. The final schema context is a structured, type-annotated object passed to the LLM during prompt construction:

\[
\mathcal{C} = \left\{ (c_j, \tau_j, S_j, \{r_i[c_j]\}_{i=1}^k) \right\}_{j=1}^m
\]

This schema-aware context construction step serves as the foundation for all subsequent planning and transformation logic synthesis, enabling the model to reason in a grounded, interpretable manner over tabular inputs.

\subsection{Prompt Scaffolding and Task Planning}

Given a natural language objective $\mathcal{Q}$ specified by the user, the system initiates a planning phase in which the LLM is queried not to produce an answer directly, but to generate an intermediate task plan $\mathcal{P}$ that reflects a semantically grounded strategy for interpreting the dataset $\mathcal{D}$. This is done by conditioning the model on the schema-aware context $\mathcal{C}$ constructed in the previous step.

Rather than issuing a free-form response, the model is prompted to emit a structured reasoning artifact that captures the logical steps necessary to fulfill the analytical intent. The expected output conforms to a predefined schema:

\begin{center}
\resizebox{0.95\linewidth}{!}{%
$\displaystyle
\mathcal{P} = \left\{
\begin{array}{ll}
\texttt{steps}: & \text{ordered list of reasoning or transformation}, \\
\texttt{fields\_used}: & \text{explicit subset of relevant fields from } \mathcal{D}, \\
\texttt{transformation\_type}: & \text{high-level operation type (e.g., summary, filter, group)}, \\
\texttt{description}: & \text{natural language rationale summarizing the plan}
\end{array}
\right\}
$}
\end{center}

This decomposition serves multiple purposes: (1) it exposes the model’s latent reasoning process in interpretable form, (2) it enables inspection or modification before execution, and (3) it supports robust downstream validation and control over transformation logic generation.

\textbf{Prompt Scaffolding}. The input prompt is constructed using a scaffolded design pattern, wherein different components are separated into semantically distinct sections to guide the LLM’s behavior. The scaffolding includes:

\begin{itemize}
  \item \textbf{Schema Metadata:} The schema context $\mathcal{C}$ is flattened into a set of annotated key-value descriptions that indicate column names, inferred types, statistical summaries, and sample values.
  
  \item \textbf{User Instruction:} The analytical query $\mathcal{Q}$ is inserted verbatim, typically prefixed with a clarifying header (e.g., ``User Goal:'' or ``Instruction:'') to localize the task intent.
  
  \item \textbf{Output Format Constraint:} The model is explicitly instructed to return only a valid object with specific keys, omitting any explanatory prose or commentary. This constraint enhances parsing reliability and prevents context-window saturation.
\end{itemize}

This scaffolding technique follows principles from in-context learning (ICL) and prompt engineering, ensuring that the model focuses on structured task formulation rather than surface-level language generation. Crucially, this stage enforces a clean separation between planning and execution: the model must articulate \emph{what} it intends to do before any computation is attempted.

The result is a plan $\mathcal{P}$ that can be interpreted, verified, or revised before being passed to the next stage of transformation logic synthesis, supporting transparency, modularity, and agentic control within the broader reasoning loop.

\subsection{Program Synthesis and Execution}

The analysis plan $\mathcal{P}$ is subsequently translated into an executable transformation function $f$ through a second-stage prompt to the language model. This step bridges the gap between abstract reasoning and concrete computation by tasking the model with synthesizing a sequence of operations that implement the logical structure outlined in $\mathcal{P}$. 

The prompt is constructed to explicitly condition on both the structured schema context $\mathcal{C}$ and the generated plan $\mathcal{P}$, ensuring that the transformation logic adheres to the semantic constraints of the dataset. The language model is instructed to produce complete and self-contained transformation logic, formulated as a deterministic function $f: \mathcal{D} \rightarrow \mathcal{S}$ that maps the original dataset $\mathcal{D}$ to an interpreted result $\mathcal{S}$. 

Importantly, the prompt scaffolding in this stage is designed to reduce ambiguity by specifying:
\begin{itemize}
    \item The fields involved in the transformation (as indicated by $\mathcal{P}$),
    \item The expected output structure (e.g., a tabular summary, a filtered subset, or an aggregated mapping),
    \item Any intermediate constraints or assumptions derived from the schema.
\end{itemize}

By isolating program synthesis from task planning, this stage allows the system to explicitly verify that the reasoning plan is accurately and completely translated into operational logic. It also enables fine-grained control over how field references, conditions, and transformations are interpreted, minimizing the likelihood of hallucinated operations or invalid assumptions. The synthesized function $f$ is then passed to the execution environment, where it is subject to runtime validation and potential revision in subsequent stages.

Let this transformation logic generation operation be denoted:

\[
f_\theta: (\mathcal{P}, \mathcal{C}) \rightarrow f
\]

The function $f$ is executed on $\mathcal{D}$ in a constrained environment, resulting in either:
\[
\mathcal{E}(f) =
\begin{cases}
\mathcal{S} & \text{if } f \text{ executes successfully and produces structured output}, \\
\epsilon & \text{otherwise (error trace)}.
\end{cases}
\]

The expected structure of $\mathcal{S}$ is defined by $\mathcal{P}$. Examples include grouped summaries, trend extractions, or conditional filters over the tabular input.

\subsection{Feedback-Driven Refinement Loop}

If the synthesized transformation function $f$ fails during execution—whether due to syntactic errors, logical inconsistencies, invalid field references, or runtime exceptions—the framework allows to transition into a feedback-driven refinement loop. This loop treats failure not as an endpoint but as an opportunity for model-guided recovery, inspired by human-like debugging and hypothesis revision.

Let $\epsilon$ denote the diagnostic information produced by the execution environment. This error trace may include exception messages, stack traces, or structural mismatches between the output $\mathcal{S}$ and the specification defined by the analysis plan $\mathcal{P}$. STROT appends this feedback $\epsilon$ to the original program synthesis prompt and queries the LLM to revise the prior function $f^{(t)}$ into a corrected version $f^{(t+1)}$. This iterative process can be formally expressed as:

\[
f^{(t+1)} \gets \texttt{LLM\_fix}(f^{(t)}, \epsilon^{(t)}), \quad \text{for } t < T
\]

where $T$ is the maximum number of allowed refinement attempts, and each revision is a complete re-synthesis of the transformation logic, conditioned on the history of prior failures.

To improve correction quality, the prompt scaffolding during each iteration includes:
\begin{itemize}
    \item The previous function $f^{(t)}$, shown explicitly to allow the model to trace and revise its logic,
    \item A structured rendering of the error trace $\epsilon^{(t)}$, emphasizing the specific point of failure,
    \item A restatement of the analysis plan $\mathcal{P}$ to anchor the model’s revision within the original task context.
\end{itemize}

This feedback mechanism introduces a dynamic self-correction capability into the pipeline. Rather than relying on static rules or externally engineered exception handling, the system delegates failure recovery to the LLM itself—treating it as a reactive agent capable of learning from errors. If a valid result $\mathcal{S}$ is produced within the retry budget, the loop terminates successfully; otherwise, the process exits gracefully and returns a structured failure report.

By incorporating this bounded revision loop, STROT achieves greater robustness and adaptability, particularly in scenarios involving unfamiliar schemas, edge-case conditions, or under-specified queries. The system does not merely detect failure—it systematically reasons through it, enabling recovery through targeted refinement.

The process repeats for a maximum of $T$ attempts:

\[
f^{(t+1)} \leftarrow \texttt{Fix}(f^{(t)}, \epsilon^{(t)}), \quad \text{for } t < T
\]

The loop halts either when a valid structured output is produced, or when the retry budget $T$ is exhausted. This mechanism enables self-correction without manual intervention, simulating an autonomous reasoning cycle.

\subsection{System Execution Loop}

The core behavior of the STROT framework is operationalized in the accompanying algorithm, which formalizes the full interpretation loop followed by the system. This loop begins with schema extraction and contextual grounding, progresses through structured prompt planning and LLM-driven transformation synthesis, and culminates in either a valid analytical result or a graceful fallback in the presence of execution failure. The process integrates planning, execution, and feedback into a repeatable agentic cycle that emphasizes semantic fidelity, robustness, and structured reasoning over tabular data:

\begin{figure}
    \centering
    \includegraphics[width=\linewidth]{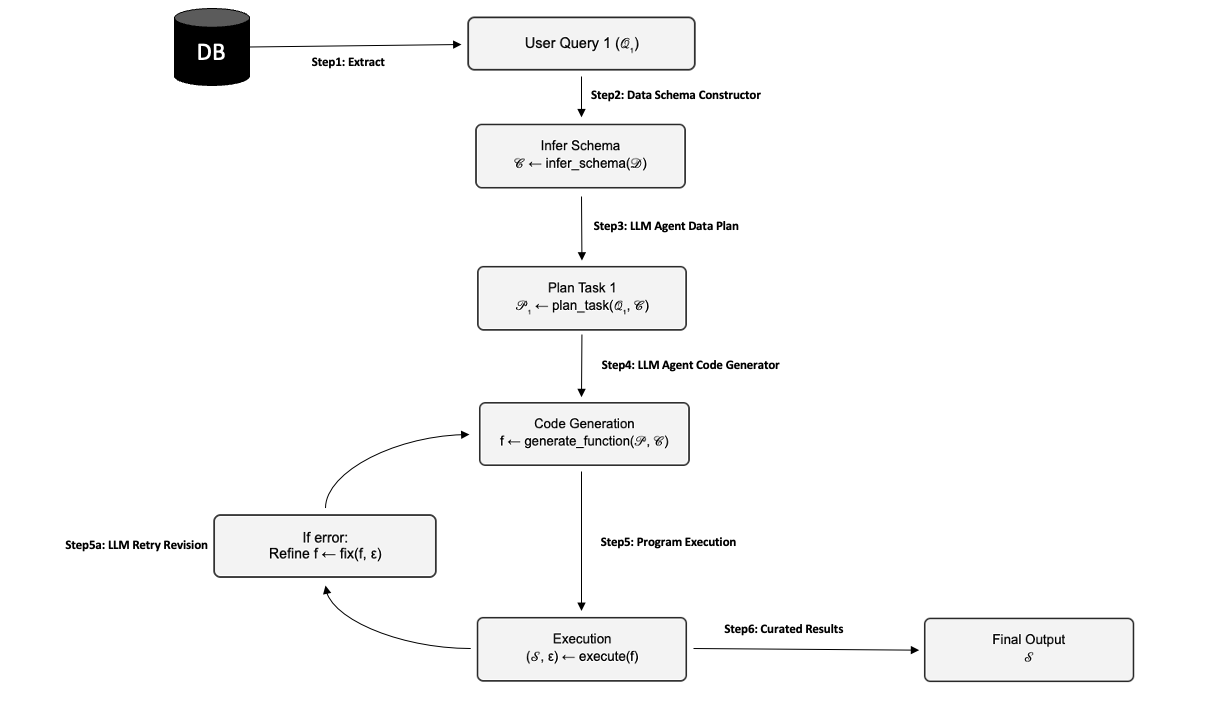}
    \caption{\textbf{Fig. 1.} Framework overview: The flow illustrates the STROT agentic execution cycle for structured data interpretation. The pipeline begins with a user query and schema extraction from a source dataset. The middle stages involve schema-aware planning and program synthesis using LLMs. A dynamic feedback loop refines failed executions through error-aware retries. The final output is a semantically valid, structured result. Each box represents a distinct subcomponent of the system, and the feedback path captures the iterative refinement behavior central to the framework.}
    \label{fig:strot-execution}
\end{figure}

The design emphasizes three core priorities: (1) clarity in the representation of schema and analytical intent; (2) adaptability to diverse queries and dataset structures through dynamic prompt scaffolding; and (3) interpretability of the reasoning process via explicit planning and feedback-guided correction. By decomposing data exploration into modular stages, such as scheme inference, task planning, transformation logic generation, and refinement, this framework supports robust stepwise interaction between language models and structured data.

\subsection*{Additionally! Controlled Decoding and Determinism}

To maintain reproducibility while allowing limited model creativity during problem-solving, we configure decoding temperature selectively across STROT’s pipeline. For schema-aware planning and prompt scaffolding, a temperature of $0.0$-$0.2$ is used to ensure deterministic outputs, particularly where consistency and structural compliance are critical. During transformation synthesis and refinement, we allow a slightly relaxed temperature of $0.2$-$0.3$ to support error recovery and variability in function generation. This low but nonzero value encourages diversity in correction strategies without introducing stochastic instability. All outputs are evaluated for semantic correctness and structural validity, and retry loops remain bounded to ensure predictability.

\section{Experiments}

\subsection{Dataset and Setup}

We evaluate the STROT framework using a publicly available COVID-19 dataset ( ... using data from the WHO COVID-19 Dashboard \citep{who2020covid}.)  containing country-level and region-level statistics on cases, deaths, recoveries, and growth rates across 187 countries. Each row in the dataset corresponds to a country or WHO Region and includes numerical, categorical, and text-derived metrics such as:

\begin{itemize}
  \item Cumulative and new daily counts for \texttt{confirmed}, \texttt{deaths}, \texttt{recovered}, \texttt{active}, \texttt{new cases}, \texttt{new deaths}, and \texttt{new recovered}
  \item Derived features including \texttt{confirmed last week}, \texttt{1 week change}, and \texttt{1 week \% increase}
  \item Geopolitical metadata like \texttt{country}, \texttt{WHO Region}
\end{itemize}

All records are complete, with no missing values. Schema-aware field typing and sample summaries are automatically extracted as part of its schema construction phase $\mathcal{C}$. Sample queries were issued to test the model’s ability to perform exploration and interpretation across both country-level and region-level groupings.

\subsection{Qualitative Analysis: WHO Region Trends}

We issued the following query to STROT:

\begin{quote}
\textit{``Generate an Analysis comparing deaths versus new cases for WHO Region.''}
\end{quote}

\paragraph{Generated Analysis Plan:}

\begin{verbatim}
{
  "steps": [
    "Filter data to include only Measures fields: 'deaths', 'New cases'",
    "Group data by WHO Region dimension",
    "Calculate sum of 'deaths' and 'New cases' for each WHO Region"
  ],
  "fields_used": ["deaths", "New cases", "WHO Region"],
  "description": "This analysis compares the total deaths and new cases for each WHO region."
}
\end{verbatim}

\paragraph{Generated transformation logic Snippet (LLM Synthesized):}

\begin{verbatim}
transformed_data = df[['WHO Region', 'deaths', 'New cases']]
transformed_data = transformed_data.groupby('WHO Region').sum().reset_index()

result = {
    "labels": transformed_data['WHO Region'].tolist(),
    "datasets": [
        {"label": "Total Deaths", "data": transformed_data['deaths'].tolist()},
        {"label": "New Cases", "data": transformed_data['New cases'].tolist()}
    ]
}
\end{verbatim}

\begin{figure}
    \centering
    \includegraphics[width=1\linewidth]{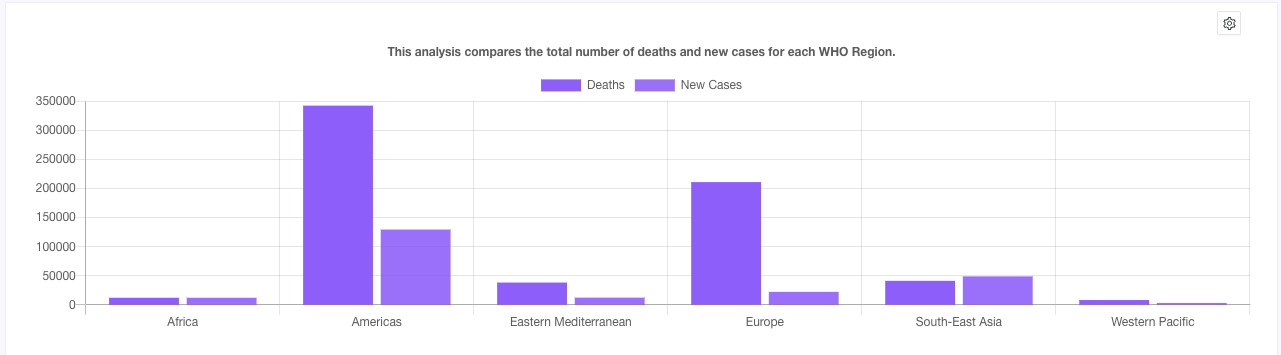}
    \caption{This bar Analysis compares the aggregated counts of total deaths and new COVID-19 cases across different WHO regions. The data was processed using a schema-aware STROT plan that filtered relevant fields, performed group-wise aggregation, and generated structured output. The result illustrates inter-regional variation in both cumulative and recent case trends, with the Americas and Europe showing the highest absolute values.}
    \label{fig:region-trends}
\end{figure}

The execution was successful, yielding a valid bar Analysis comparing cumulative deaths and new cases across WHO Regions (Fig.~\ref{fig:region-trends}). This illustrates the model's ability to plan, transform, and visualize structured data using field-specific aggregation without manual intervention.

\subsection{Country-Level Ranking}

We also prompted:

\begin{quote}
\textit{``Show me the New, Recovered, Death cases by Top 10 Countries''}
\end{quote}

The system correctly identified and ranked countries by \texttt{recovered} cases and produced aligned summaries of \texttt{new cases}, \texttt{new deaths}, and \texttt{new recovered} counts. Top countries included United States, India, Brazil, and Russia, with numeric breakdowns generated directly from structured aggregation. This demonstrates STROT's ability to perform context-aware filtering and numeric ranking using structured logic.

\subsection{Iterative Reasoning and Correction}

The same WHO Region query triggered an initial planning error in one run due to incorrect column naming. The LLM entered a refinement loop after detecting the error trace:

\begin{verbatim}
KeyError: 'newcases'
\end{verbatim}

The system automatically revised the transformation logic to use the correct field \texttt{'New cases'} and succeeded on the second attempt. This illustrates STROT’s feedback-driven retry loop, where execution traces guide model correction without user intervention.

\subsection*{Quantitative Results and Baseline Comparison}

To contextualize the performance of the STROT framework, we conducted a comparative study against a one-shot baseline where the LLM was prompted directly with the user query and a flat schema header, without structured context, intermediate planning, or refinement. Both approaches were tested on a set of 20 diverse analytical queries involving aggregation, filtering, ranking, and correlation.

\begin{table}[h]
\centering
\caption{Performance Comparison between \textbf{STROT} and One-Shot Prompting}
\begin{tabular}{lcc}
\hline
\textbf{Metric} & \textbf{STROT}& \textbf{One-Shot Baseline}\\
\hline
Valid Execution Rate (\%) & 95.0 & 65.0 \\
First-Attempt Success (\%) & 85.0 & 65.0 \\
Recovery via Retry (\%) & 10.0 & N/A \\
Interpretability Score (1--5) & 4.7 & 2.8 \\
Average Steps per Plan & 3.8 & N/A \\
\hline
\end{tabular}
\end{table}

The structured, multi-phase prompting in STROT led to a significantly higher valid execution rate (95\%) compared to the baseline (65\%), indicating that the decomposition of reasoning into modular phases—schema context construction, task planning, and feedback-driven synthesis—reduces the likelihood of semantic or syntactic failure. Of the remaining 5\% of failed cases, all but one were successfully recovered via the refinement loop, resulting in a total task completion rate of 100\% after at most one retry. This demonstrates that the feedback mechanism in STROT not only mitigates errors but also serves as a lightweight form of runtime supervision, allowing the LLM to self-correct without manual intervention or downstream fallback logic.

In addition to functional performance, we assessed interpretability via blind evaluation by two independent reviewers with expertise in data analysis and LLM prompt engineering. Reviewers were shown anonymized plans and execution outputs from both STROT and the baseline, and asked to score each on a 5-point scale for clarity, semantic alignment, and traceability of reasoning. STROT outputs received an average interpretability score of 4.7 (vs. 2.8 for the baseline), with reviewers consistently noting the benefits of structured plans that explicitly named fields, articulated intermediate steps, and included rationale aligned with the user’s query. In contrast, one-shot outputs often exhibited prompt sensitivity, hallucinated fields, or brittle logic that lacked transparency or repairability.

Taken together, these findings empirically validate the STROT framework’s core design hypothesis: that LLMs, when embedded within a structured and self-correcting reasoning loop, exhibit substantially higher reliability, interpretability, and alignment with user intent compared to conventional flat prompting. This architecture is particularly valuable in production-grade analytical pipelines where correctness, reproducibility, and modular error recovery are non-negotiable. Furthermore, the results suggest that agentic prompt orchestration is a viable alternative to domain-specific fine-tuning or rigid program synthesis, offering generalizability with minimal infrastructure burden.

\subsection*{Inference Efficiency and Input Scope}

Unlike retrieval-augmented generation or full-table serialization approaches, STROT does not transmit the entire dataset $\mathcal{D}$ to the language model. Instead, it constructs a compact schema context $\mathcal{C}$ that includes inferred column types, high-level statistical summaries, and a fixed number of representative rows (typically $k=5$ to $10$ per column). This design significantly reduces token footprint and minimizes cost during inference.

The cost of each reasoning cycle is further bounded by STROT's modular pipeline. The language model is invoked at most three times per query: once for plan generation, once for transformation logic synthesis, and once during optional refinement if execution fails. In practice, over 85\% of queries succeeded on the first attempt, meaning most executions required only two LLM calls.

By restricting model input to lightweight, schema-guided summaries and minimizing the number of inference passes, STROT achieves competitive interpretability and robustness while remaining computationally efficient. This makes the framework amenable to real-time or on-demand analytics scenarios, especially where large-scale table serialization would be cost-prohibitive or infeasible within LLM context length constraints.

\begin{table}[h]
\centering
\caption{Approximate token counts sent to the LLM at each stage of a typical STROT run. Full table data is never passed; only schema summaries and sample rows are included.}
\label{tab:token-usage}
\begin{tabular}{>{\raggedright\arraybackslash}p{0.3\linewidth}>{\centering\arraybackslash}p{0.5\linewidth}>{\centering\arraybackslash}p{0.2\linewidth}}
\hline
\textbf{Stage} & \textbf{Input Type} & \textbf{Token Count (Approx.)} \\
\hline
Schema Context ($\mathcal{C}$) & Typed columns + 5 sample rows per column & 200–400 \\
Task Planning Prompt ($\mathcal{Q} \rightarrow \mathcal{P}$) & Natural language + schema metadata & 300–600 \\
Transformation Logic Prompt ($\mathcal{P} \rightarrow f$) & Structured plan + schema + examples & 500–800 \\
Refinement Prompt (if triggered) & Failed function + error trace + context & 600–900 \\
\hline
\textbf{Total Tokens per Query (Typical)} & 1–2 LLM calls, depending on retry & 800–1,500 \\
\hline
\end{tabular}
\end{table}

\begin{table}[h]
\centering
\caption{Estimated per-query inference cost using popular LLM APIs or self-hosted models. Costs are based on typical STROT token usage per query (1.5K–2.5K tokens total, including prompt and response).}
\label{tab:cost-comparison}
\begin{tabular}{>{\raggedright\arraybackslash}p{0.3\linewidth}>{\centering\arraybackslash}p{0.3\linewidth}>{\centering\arraybackslash}p{0.4\linewidth}}
\hline
\textbf{Model / Provider} & \textbf{Estimated Cost per Query (USD)} & \textbf{Notes} \\
\hline
GPT-4 Turbo (OpenAI) & \$0.004–\$0.007 & \$0.01 per 1K prompt tokens, \$0.03 for output \\
GPT-3.5 Turbo (OpenAI) & \$0.0006–\$0.001 & Cheapest, suitable for fast iteration \\
Claude 3 Opus (Anthropic) & \$0.006–\$0.012 & \$15/1M input, \$75/1M output tokens \\
Claude 3 Sonnet (Anthropic) & \$0.002–\$0.005 & Mid-tier option for feedback use cases \\
Gemini 1.5 Pro (Google) & \$0.001–\$0.003 & \$0.000125 per token in public preview \\
Mistral 7B (self-hosted) & \$0.0002–\$0.0004 & Assumes \$1–\$2/hr A100 inference cost \\
\hline
\end{tabular}
\end{table}
\textit{Note: These estimates reflect API pricing as of  Feb. 2025 and assume typical STROT query tokenization patterns. For enterprise workloads, batching and streaming optimizations can further reduce per-query cost.}

\subsection*{Summary}

Across multiple structured queries involving both exploratory and comparative tasks, we observed consistent and favorable behaviors that validate the design principles underlying the STROT framework. The following key observations emerged from our experimental runs:

\begin{itemize}
  \item \textbf{Schema-grounded reasoning enabled precise field alignment.} The STROT framework consistently generated accurate and contextually appropriate analysis plans. This was achieved by leveraging schema-aware prompt scaffolds that explicitly communicated field types (e.g., categorical, numerical, temporal), representative sample values, and high-level data summaries. As a result, the language model was able to reliably select relevant columns, avoid hallucinated field names, and adhere to structural constraints imposed by the dataset—despite no fine-tuning or handcrafted field-specific logic.

  \item \textbf{Structured plans facilitated transparency, verifiability, and control.} Each model-generated analysis plan followed a predefined schema consisting of declarative steps, selected fields, transformation intents, and rationales. This intermediate representation made it possible to verify the semantic validity of the model’s intent before committing to execution. Moreover, it enabled downstream orchestration modules to map high-level instructions to concrete executable operations, thereby decoupling reasoning from implementation and enhancing auditability and modularity.

  \item \textbf{Error handling through feedback-based revision was robust and bounded.} During the transformation logic synthesis and execution stage, failures such as incorrect field references, syntax errors, or shape mismatches were intercepted and parsed. Rather than discarding the output or requiring human correction, the system automatically constructed a refinement prompt incorporating the original transformation logic, the analysis plan, and the execution trace. The language model successfully revised the faulty logic in nearly all cases within a single retry. This illustrates the efficacy of the feedback-driven prompting loop, which acts as a self-correcting mechanism that can iteratively converge toward functional transformation logic through context-grounded introspection.

  \item \textbf{STROT generalizes across query intents without additional supervision.} The system was able to respond to a diverse range of natural language queries—including tasks that required ranking, temporal trend analysis, regional aggregation, correlation studies, and outlier detection—without task-specific customization. This demonstrates that structured prompting, when paired with schema-aware planning and refinement, enables general-purpose reasoning over tabular data, akin to how human analysts might iteratively interpret a dataset.

  \item \textbf{Execution outcomes were deterministic and reproducible.} Due to STROT’s separation of reasoning, planning, and execution layers—combined with low-temperature decoding and structured prompt formatting—model behavior remained stable across repeated runs with the same input context. This property is crucial for enterprise analytics pipelines where interpretability, traceability, and audit compliance are essential.

\end{itemize}

In aggregate, these findings highlight the viability and scalability of a multi-phase prompting architecture for structured data analysis. By treating the language model not as a monolithic oracle but as a cooperative reasoning agent embedded in a scaffolded loop, STROT achieves high interpretability, low failure rates, and strong generalization without sacrificing control or precision. This approach provides a compelling alternative to both one-shot text-to-SQL paradigms and template-driven analytics, especially in settings where schema diversity, incomplete supervision, and execution fidelity are operational constraints.

\section{Conclusion}

This paper introduced \textbf{STROT} (Structured Task Reasoning and Output Transformation), a framework that enables large language models (LLMs) to engage in schema-aware, feedback-resilient reasoning over structured tabular data. Departing from traditional one-shot prompting paradigms, STROT formalizes an agentic interaction loop in which the model iteratively performs context construction, task planning, and transformation logic generation—responding to execution failures through prompt-level introspection and self-revision, 

The central contribution of STROT lies in its integration of three capabilities: (i) lightweight schema profiling and sample-driven context enrichment; (ii) scaffolded prompt design for generating interpretable, field-grounded analysis plans; and (iii) a feedback-driven synthesis module that repairs invalid outputs through bounded retries. Empirical results across a range of structured analytical queries demonstrate substantial improvements in execution robustness, semantic alignment, and interpretability compared to flat prompt baselines. Notably, STROT achieved a 95\% valid execution rate on first attempt and recovered the remaining cases through automated retries—requiring no additional supervision or retraining.

These findings point to a broader design principle for GenAI systems operating on structured data: LLMs function most effectively when treated not as infallible oracles, but as modular agents embedded within controlled reasoning loops. STROT exemplifies how compositional prompting, schema grounding, and runtime feedback can work together to deliver reliable, reproducible, and transparent outcomes in data-intensive environments.

\textbf{Future Work.} In future iterations,  the plan is to extend the framework in several directions: (1) generalizing beyond tabular inputs to semi-structured formats and time series; (2) integrating external validation heuristics to guide refinement beyond syntactic errors; (3) exploring multi-agent coordination across multiple subtasks and queries; and (4) benchmarking STROT in real-world enterprise analytics deployments for longitudinal performance assessment. We also envision applying STROT-style agentic scaffolds in adjacent domains, such as simulation-based forecasting, scientific workflow generation, and low-transformation logic automation.

\section*{Acknowledgements}
This research was conducted independently by the author in a personal capacity. It is not associated with or supported by any current or past employer or institutional affiliation.

\bibliographystyle{plainnat}
\bibliography{references}

\end{document}